\documentclass[sigconf]{acmart}

\usepackage{booktabs} 
\usepackage{subfigure}
\usepackage{multirow}
\usepackage{flushend}

\begin{document}
\title{Learning and Transferring IDs Representation in E-commerce}

\author{Kui Zhao, Yuechuan Li, Zhaoqian Shuai, Cheng Yang}
\affiliation{
  \institution{Machine Intelligence Technologies, Alibaba Group}
  \city{Hangzhou} 
  \state{Zhejiang} 
  \country{China}
  \postcode{311121}
}
\email{{zhaokui.zk, yuechuan.lyc, wanqian.szq, charis.yangc}@alibaba-inc.com}

\begin{abstract}
Many machine intelligence techniques are developed in E-commerce and one of the most essential components is the representation of IDs, including user ID, item ID, product ID, store ID, brand ID, category ID etc. The classical encoding based methods (like one-hot encoding) are inefficient in that it suffers sparsity problems due to its high dimension, and it cannot reflect the relationships among IDs, either homogeneous or heterogeneous ones. In this paper, we propose an embedding based framework to learn and transfer the representation of IDs. As the implicit feedbacks of users, a tremendous amount of item ID sequences can be easily collected from the interactive sessions. By jointly using these informative sequences and the structural connections among IDs, all types of IDs can be embedded into one low-dimensional semantic space. Subsequently, the learned representations are utilized and transferred in four scenarios: (i) measuring the similarity between items, (ii) transferring from seen items to unseen items, (iii) transferring across different domains, (iv) transferring across different tasks. We deploy and evaluate the proposed approach in Hema App and the results validate its effectiveness. 
\end{abstract}

\copyrightyear{2018} 
\acmYear{2018} 
\setcopyright{acmcopyright}
\acmConference[KDD '18]{The 24th ACM SIGKDD International Conference on Knowledge Discovery \& Data Mining}{August 19--23, 2018}{London, United Kingdom}
\acmBooktitle{KDD '18: The 24th ACM SIGKDD International Conference on Knowledge Discovery \& Data Mining, August 19--23, 2018, London, United Kingdom}
\acmPrice{15.00}
\acmDOI{10.1145/3219819.3219855}
\acmISBN{978-1-4503-5552-0/18/08}

%
%



\maketitle

\section{Introduction}
\label{section:intro}
E-commerce has become an important part of our daily lives and the increasing online shopping 
makes it a well-known concept. However, we still have little knowledge about E-commerce 
because it has much more dynamic and complex business environment than traditional commerce. 
Fortunately, many data can be tracked and collected from multiple sources in E-commerce 
and then machine intelligence techniques can help us to make more informed decisions 
by analyzing and modeling these data.  
The intelligence techniques allow us to gain deeper insights into customer behaviors and industry 
trends, and subsequently make it possible to improve many aspects of the business, including marketing, 
advertising, operations, and even customer retention etc. 
One of the essential elements of machine intelligence techniques is the representation of 
data \cite{bengio2013representation}. It is the fundamental step for intelligent techniques to understand our world. 
In E-commerce, one important type of data is the unordered discrete data and we usually call them IDs, 
including user ID, item ID, product ID, store ID, brand ID and category ID etc. 

The traditional technique treats IDs as atomic units and represents them as indices or one-hot encodings. 
The simplicity and robustness make it very popular in statistical models. However, it has two main limitations. 
Firstly, it suffers from sparsity problems due to the high dimension. 
The size of the space spanned by $N$ different IDs is $2^N$. That means the samples we need to make the 
statistical models have enough confidence increase exponentially as the number of IDs increases. 
Secondly, it cannot reflect the relationships 
among IDs, either homogeneous or heterogeneous ones. Taking two different item IDs as the homogeneous example, 
two item IDs have a constant distance measured by one-hot encodings no matter whether they are similar or not. 
Taking the item ID and store ID as the heterogeneous example, the relationship between them even cannot be measured 
since they are in different spaces.

Recently, word embedding (or called word2vec) techniques \cite{mikolov2013efficient, mikolov2013distributed}, 
which are originated from the Natural Language Processing (NLP) domain, have attracted a considerable amount 
of interests from various domains. These techniques embed words into the low-dimensional representation to 
capture syntactical and semantic relationships among words. 
In E-commerce, the item ID is the core interactive unit and a tremendous amount of sequences on it 
can be easily collected from the interactive sessions, which are the implicit feedbacks of users. 
Inspired by word2vec, item2vec \cite{barkan2016item2vec} embeds item IDs into the low-dimensional 
representation by modeling the item ID co-occurrence in interactive sequences. The item2vec techniques can be 
used to boost the accuracy of recommendation \cite{sun2017mrlr} and search \cite{ai2017learning} in E-commerce. 

\begin{figure}[htbp!]
\centering
\includegraphics[width=7cm]{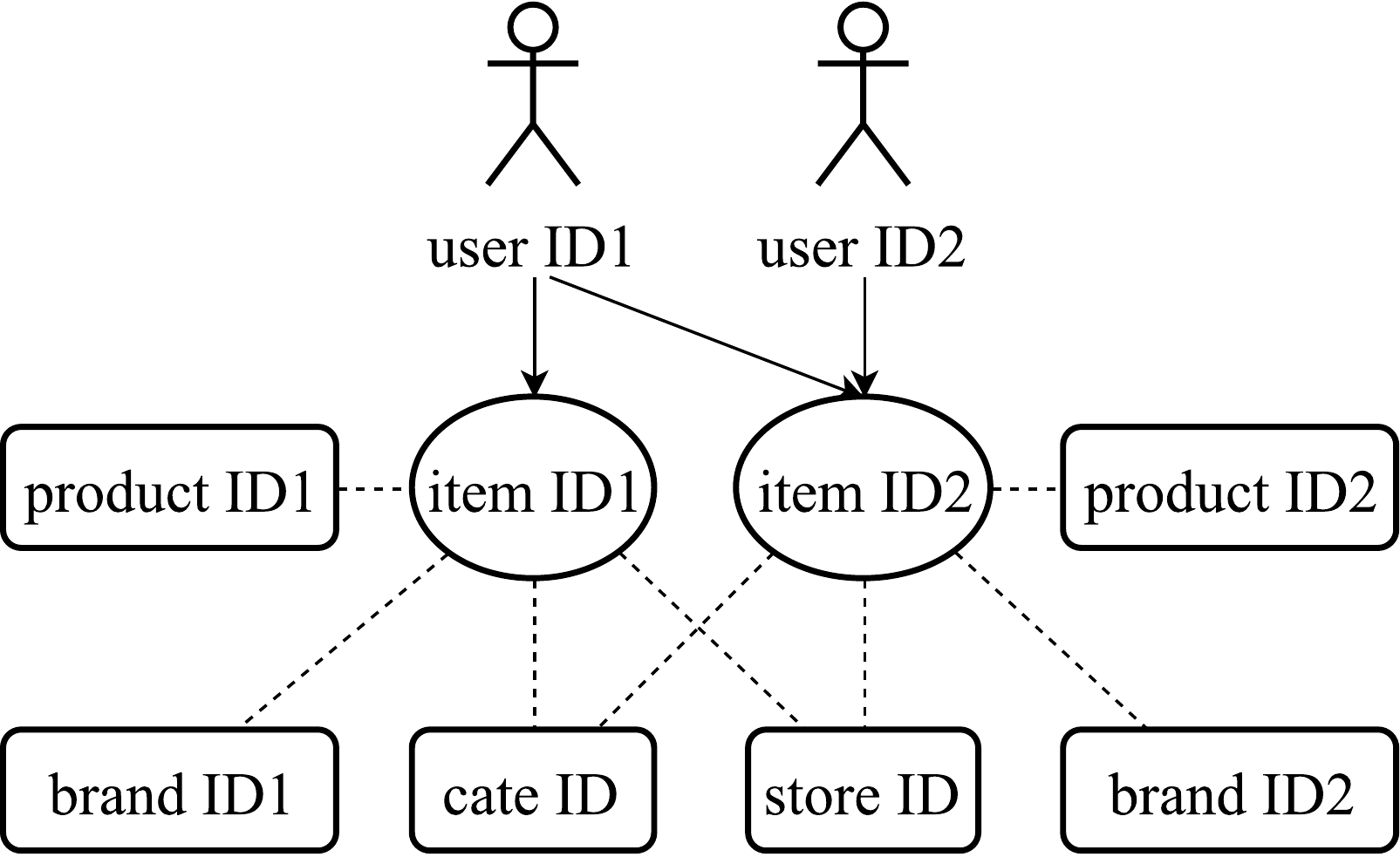}
\caption{The item ID and its structural connections with other types of IDs.}
\label{fig:ids}
\end{figure}

In this paper, we extend the item2vec and present an embedding based framework to 
learn and transfer low-dimensional representations for all types of IDs. 
Besides the implicit feedbacks of users, we also consider the structural connections 
between item ID and other types of IDs (as illustrated in Figure \ref{fig:ids}). 
Through these connections, the information indicated in item ID sequences can propagate to other types of IDs, 
and the representations of all types of IDs can be learned simultaneously. 
In the proposed framework, all types of IDs are embedded into one low-dimensional semantic space, 
where the relationships among IDs, both homogeneous and heterogeneous ones, can be measured conveniently. 
This property makes it very easy to deploy and transfer the learned representations in real world scenarios. 
With the assist of transfer learning \cite{pan2010survey}, 
the information contained in interactive sequences on item IDs can be leveraged in many applications. 
We deploy and evaluate the proposed approach in Hema App and the results validate its effectiveness: 
\begin{itemize}
\item {\it Measuring the similarity between items.} 
Computing item-item relationships is a key 
building block in modern recommendation systems. We will see that measuring item similarities by 
cosine distances among embedding vectors has a higher recall score than the classical item-based 
Collaborative Filtering (CF) algorithm.  

\item {\it Transferring from seen items to unseen items.} 
New items unavoidably cause cold-start problem, 
which means item IDs with no historical records are invisible to the recommendation systems. 
In our work, approximate embedding vectors are constructed for new item IDs 
by transferring the embedding vectors of seen IDs. 
The experimental results show that the constructed vectors are very encouraging. 

\item {\it Transferring across different domains.} 
For emerging platforms like Hema, a high proportion of users are new customers, 
and thus the personalized recommendation is very challenging. 
In our approach, embedding vectors of user IDs are constructed by aggregating embedding vectors of item IDs. 
We will see that these vectors can be transferred from long existing platforms like Taobao, 
onto the emerging platforms to provide effective personalized recommendation for new users. 

\item {\it Transferring across different tasks.} Sales forecast is very helpful for making informed business decisions. 
When forecasting the sales for different stores, it is important to properly represent store IDs. 
We will see that the embedding vectors of store IDs learned by our approach 
can be transferred to this new task and they are more effective than the one-hot encoding. 
\end{itemize}

The rest of our paper is organized as follows. 
The service process of Hema App is introduced in Section 2. 
We describe how to learn the IDs representation in Section 3, 
and how to deploy and transfer the IDs representation in Section 4. 
We present the experimental setup and results in Section 5. 
We briefly review related work in Section 6. 
Finally, the conclusions and future plans are given in Section 7. 

\section{The Service Process of Hema}
\label{se:hema}
\begin{figure}[!htbp]
\centering
\subfigure[]{
\label{fig:hema:a} 
\begin{minipage}[t]{0.4\linewidth}
\includegraphics[width=1.\textwidth]{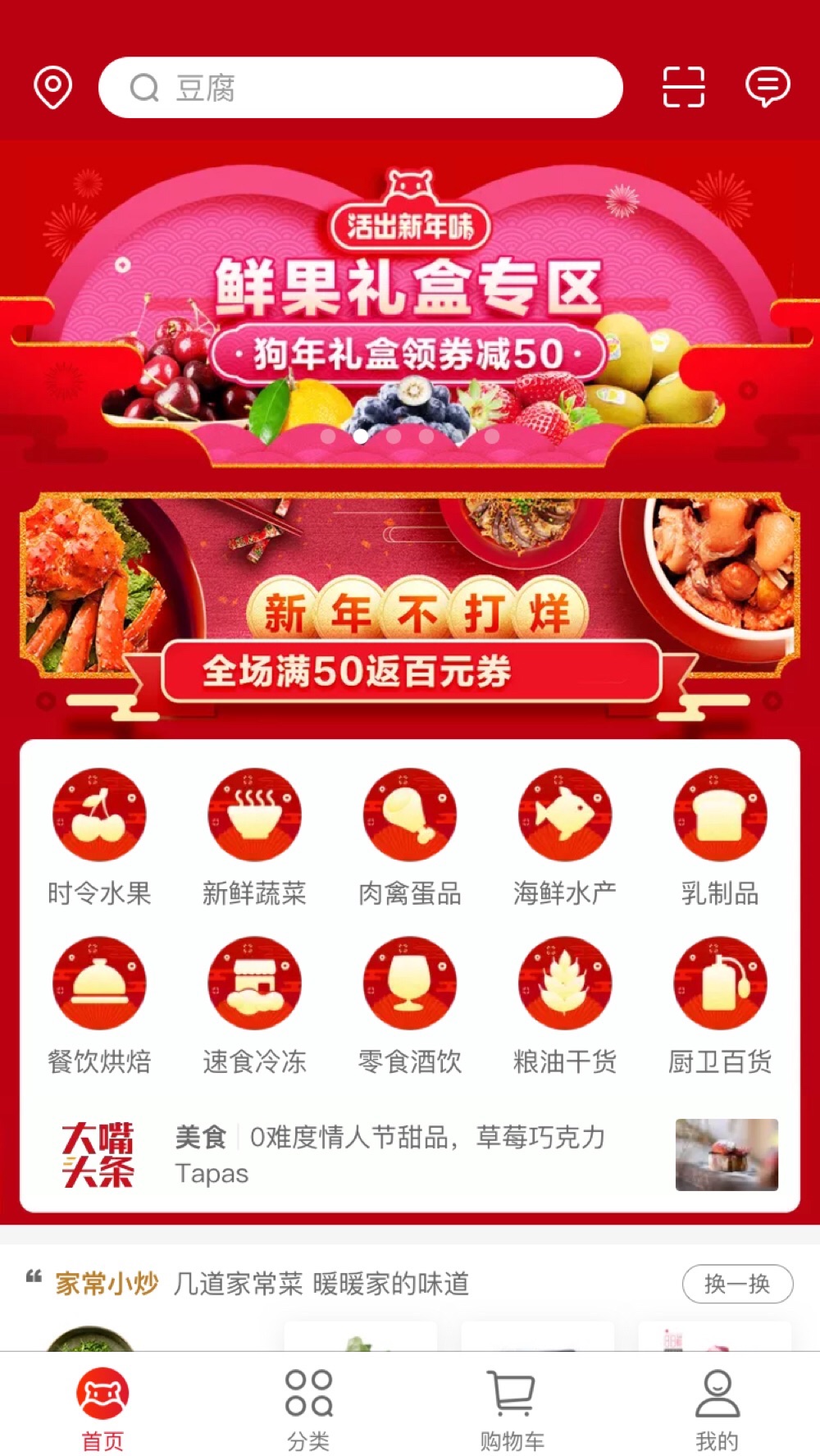}
\end{minipage}
}
\hspace{0.0in}
\subfigure[]{
\label{fig:hema:b} 
\begin{minipage}[t]{0.4\linewidth}
\includegraphics[width=1.\textwidth]{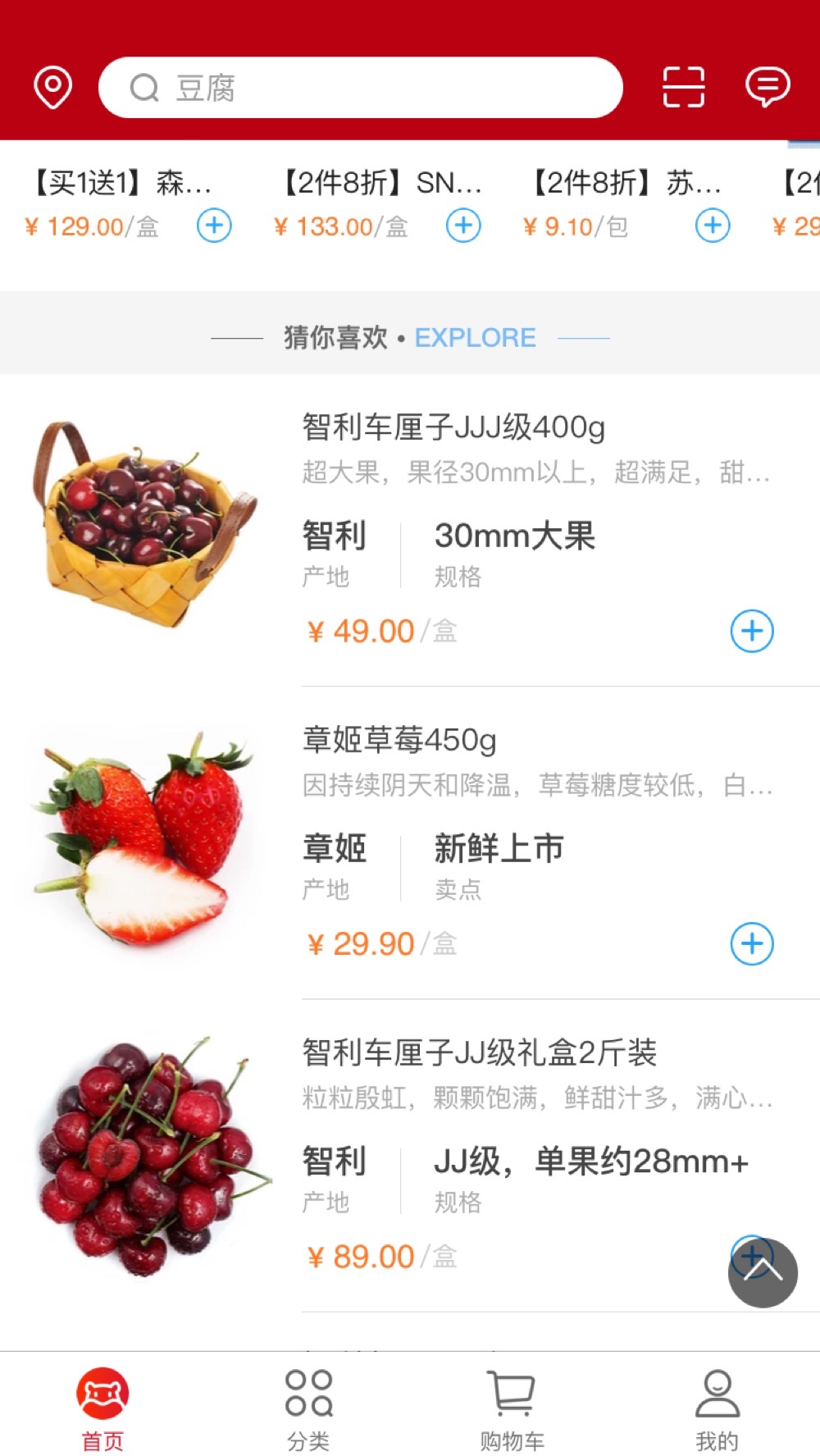}
\end{minipage}
}
\caption{An example of homepage on Hema App.}
\label{fig:hema} 
\end{figure}

Hema App is an online-to-offline (O2O) service platform provided by Alibaba Group. 
As shown in Figure \ref{fig:hema}, customers can either browse items by category 
or explore items in recommendation lists. 
Unlike other online stores, Hema mainly focuses on fresh food, 
where items are updated seasonally and thus many unseen items arise frequently. 
Besides, Hema is an emerging platforms and a large proportion of users are new customers. 
These factors make the personalized recommendation very challenging on it. 
Moreover, it is very important to deliver the package in time due to the fresh characteristic. 
So we need forecasting the delivery demand to pre-order appropriate number of delivery staff. 

In conventional recommendation framework, the Click Through Rates (CTR) are calculated between users and items. 
However, it is impossible to calculate and store CTR scores between all users and all items 
since there are too many user-item pairs (usually in size of $10^{15}-10^{20}$). 
In our work, the recommendation framework is divided into four processes to overcome this challenge: 
\begin{itemize}
\item {\bf Preparation}. The user-to-trigger (u2t) preference scores and the trigger-to-item (t2i) matching scores are calculated offline, 
and the results are stored in key-key-value databases for efficient online retrieving.  
\item {\bf Matching}. For each user access, triggers are firstly retrieved according to the user ID. 
Then a candidate set of recommended items is obtained based on these tiggers.  
\item {\bf Filtering}. Remove duplicated and invalid items, such as sold-out items etc. 
\item {\bf Ranking}. Rank filtered items according to an integrated score, which considers preference score, matching score, 
and other business objectives. 
\end{itemize}
 
The size of candidate set and the amount of computation can be controlled by setting the number of triggers, 
which makes the entire framework much more flexible. 
In practice, various triggers are used in the system, such as visited items, categories and so on. 

\section{Learning IDs Representation}
In this section, we discuss how to embed all types of IDs in E-commerce into one semantic space. 
We explore both users' interactive sequences on item IDs and 
the structural connections among different types of IDs. 
\subsection{Skip-gram on User's Interactive Sequences}
\begin{figure}[ht!]
\centering
\includegraphics[width=7.5cm]{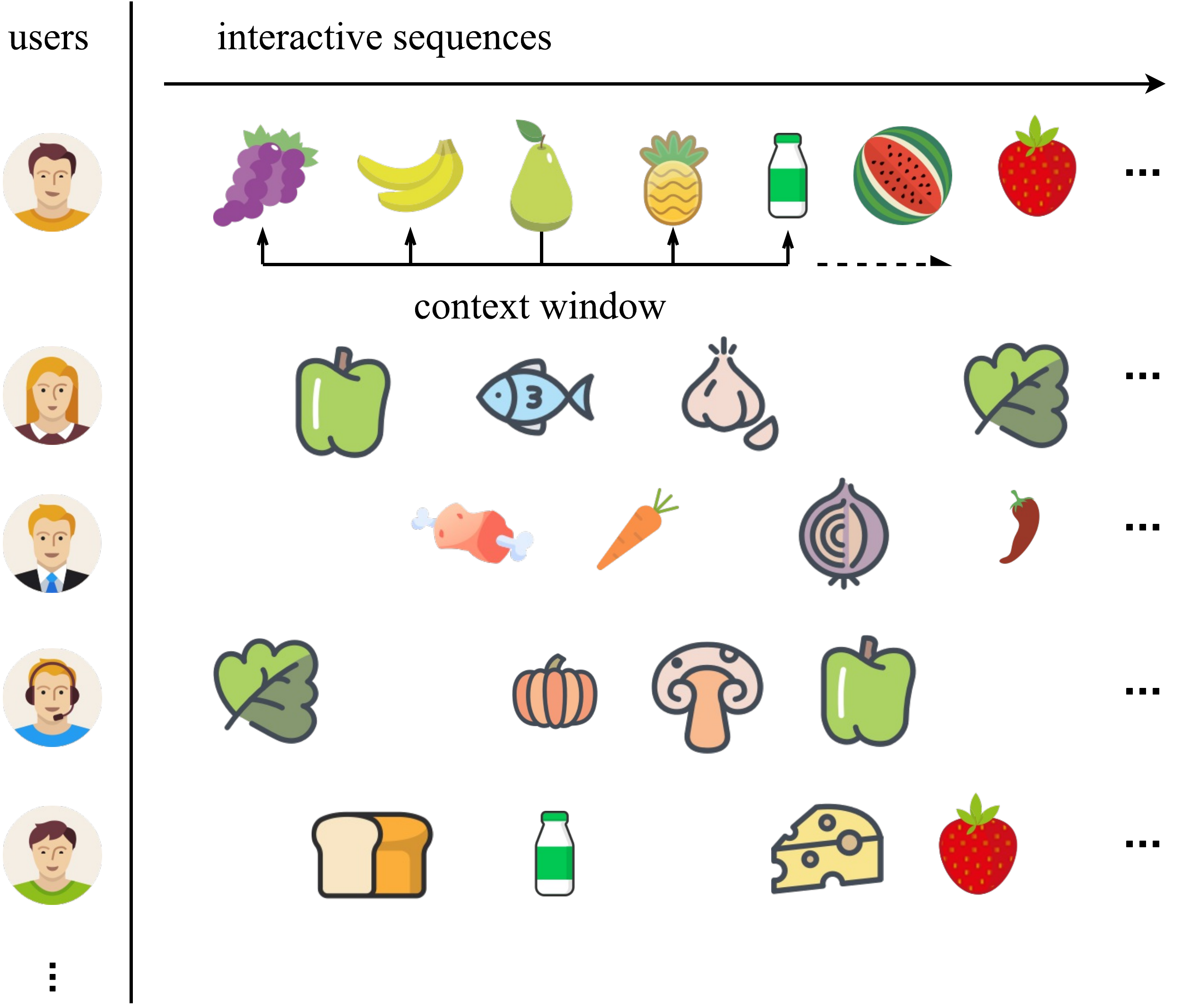}
\caption{The skip-gram model with context window of length $C=2$.}
\label{fig:skip_gram}
\end{figure}

In E-commerce, item is the core interactive unit. We can obtain massive item ID sequences 
from different interactive sessions, which are the implicit feedback of users. 
By regarding them as ``documents'', skip-gram model is to find the representations that are useful for predicting the 
surrounding item IDs given the target item ID in a sequence, which is illustrated in Figure \ref{fig:skip_gram}. 
More formally, given a sequence of item IDs $\{\text{item}_1, \dots, \text{item}_i, \dots, \text{item}_N\}$, 
the objective of the skip-gram model is to maximize the average log probability:
\begin{equation}
\mathcal{J}=\frac{1}{N}\sum\limits_{n=1}^{N}\sum\limits_{-C\le j \le C}^{1 \le n+j \le N, j \ne 0}\log p(\text{item}_{n+j}|\text{item}_n),
\label{eq:lp}
\end{equation}
where $C$ is the length of context window. 

In the basic skip-gram model, $p(\text{item}_j|\text{item}_i)$ is defined by the softmax function:
\begin{equation}
p(\text{item}_j|\text{item}_i)=\frac{\exp({{\bf e}'_j}^{\rm T}{\bf e}_i)}{\sum_{d=1}^{D}\exp({{\bf e}'_d}^{\rm T}{\bf e}_i)},
\end{equation}
where ${\bf e}'\in {\bf E}'(\subset \mathbb{R}^{m\times D})$ and ${\bf e}\in {\bf E}(\subset \mathbb{R}^{m\times D})$. 
${\bf E}'$ and ${\bf E}$ are matrices corresponding to the context and target representations respectively. 
$m$ is the dimension of embedding vectors. 
 $D$ is the size of the dictionary that contains all item IDs. 
 
\subsection{Log-uniform Negative-sampling}
The computational cost of $\nabla \log p(\text{item}_j|\text{item}_i)$ is proportional to $D$, 
which makes it impractical when $D$ is very large. 
Noise Contrastive Estimation (NCE) assumes that a good model is able to differentiate data from noise
and subsequently approximates the log probability of softmax \cite{gutmann2012noise, mnih2012a} 
with the means of logistic regression. 
While the skip-gram model is only concerned with learning high-quality vector representation, 
Mikolov et al. simplify NCE by negative-sampling \cite{mikolov2013distributed} and replace softmax function with
\begin{equation}
p(\text{item}_j|\text{item}_i)=\sigma({{\bf e}'_j}^{\rm T}{\bf e}_i)\prod\limits_{s=1}^{S}\sigma(-{{\bf e}'_s}^{\rm T}{\bf e}_i),
\label{eq:old_match}
\end{equation} 
where $\sigma(x)=\frac{1}{1+\exp(-x)}$ is sigmoid function and 
$S$ is the number of negative samples to be drawn from the noise distribution 
$P_\text{neg}(\text{item})$ for each positive sample.
While NCE requires both the sampling and the numerical probability of noise distribution, 
negative-sampling only needs the sampling. 

The noise distribution $P_\text{neg}(\text{item})$ is a free parameter and the simplest choice is the uniform distribution. 
However, the uniform distribution cannot count the imbalance between rare and frequent items. 
The popular item IDs appear in context window of many target item IDs and such item IDs provide little information.
In this paper, we balance them with the log-uniform negative-sampling, where negative samples are drawn from the 
Zipfian distribution \cite{powers1998applications}. 
Zipfian distribution approximates many types of data in the physical and social sciences, 
and items subject to it quite well. 
That means the appearing frequency of any item is inversely proportional to its rank in the frequency table. 
To further speed up the negative-sampling procedure, we first sort items in decreasing order of frequency and 
index them in range $[0,  D)$ according to their ranks. 
Then we can approximate Zipfian distribution with 
\begin{equation}
p(\text{index})=\frac{\log(\text{index}+2)-\log(\text{index}+1)}{\log(D+1)}.
\end{equation}
The cumulative distribution function is: 
\begin{equation}
\begin{aligned}
F(x)&=p(x\le index) \\
& = \sum\limits_{i=0}^\text{index}\frac{\log(\text{i}+2)-\log(\text{i}+1)}{\log(D+1)}\\
& = \frac{\log(\text{index}+2)}{\log(D+1)}.
\end{aligned}
\end{equation}
Let $F(x)=r$ and $r$ is a random number drawn from the uniform distribution $U(0, 1]$. 
The sampling of Zipfian distribution can be approximated by: 
\begin{equation}
\text{index} = \lceil(D+1)^r\rceil - 2. 
\end{equation}
By utilizing the above approximation, the log-uniform negative-sampling is very computationally efficient. 

\subsection{IDs and Their Structural Connections}
There are many types of IDs (as illustrated in Figure \ref{fig:ids}) and they can be divided into two groups: 
\subsubsection{Item ID and its attribute IDs} 
\label{se:ia}
Item is the core interactive unit in E-commerce and it has many attribute IDs, 
including product ID, store ID, brand ID and category ID etc. 
We here give a brief illustration about them.
The ``product'' is a basic concept, e.g., 
``iPhone X'' sold in different stores shares the same {\it product ID}. 
When one product is sold in several stores, it has a specific {\it item ID} in each store (indicated by {\it store ID}). 
Each product belongs to some brand (indicated by {\it brand ID}) and some categories (indicated by {\it category IDs}). 
It is noteworthy that category IDs have many levels, such as {\it cate-level1 ID}, {\it cate-level2 ID} 
and {\it cate-level3 ID} etc. Although many other attributes IDs can be considered, 
we focus on seven types of IDs in this paper without loss of generality. 
They are item ID, product ID, store ID, brand ID, cate-level1 ID, cate-level2 ID and cate-level3 ID.

\subsubsection{User ID} 
A user can be identified by the user ID, such as cookie, device IMEI, or log-in username etc. 

\subsection{Jointly Embedding Attribute IDs}
\begin{figure}[htbp!]
\centering
\includegraphics[width=7.5cm]{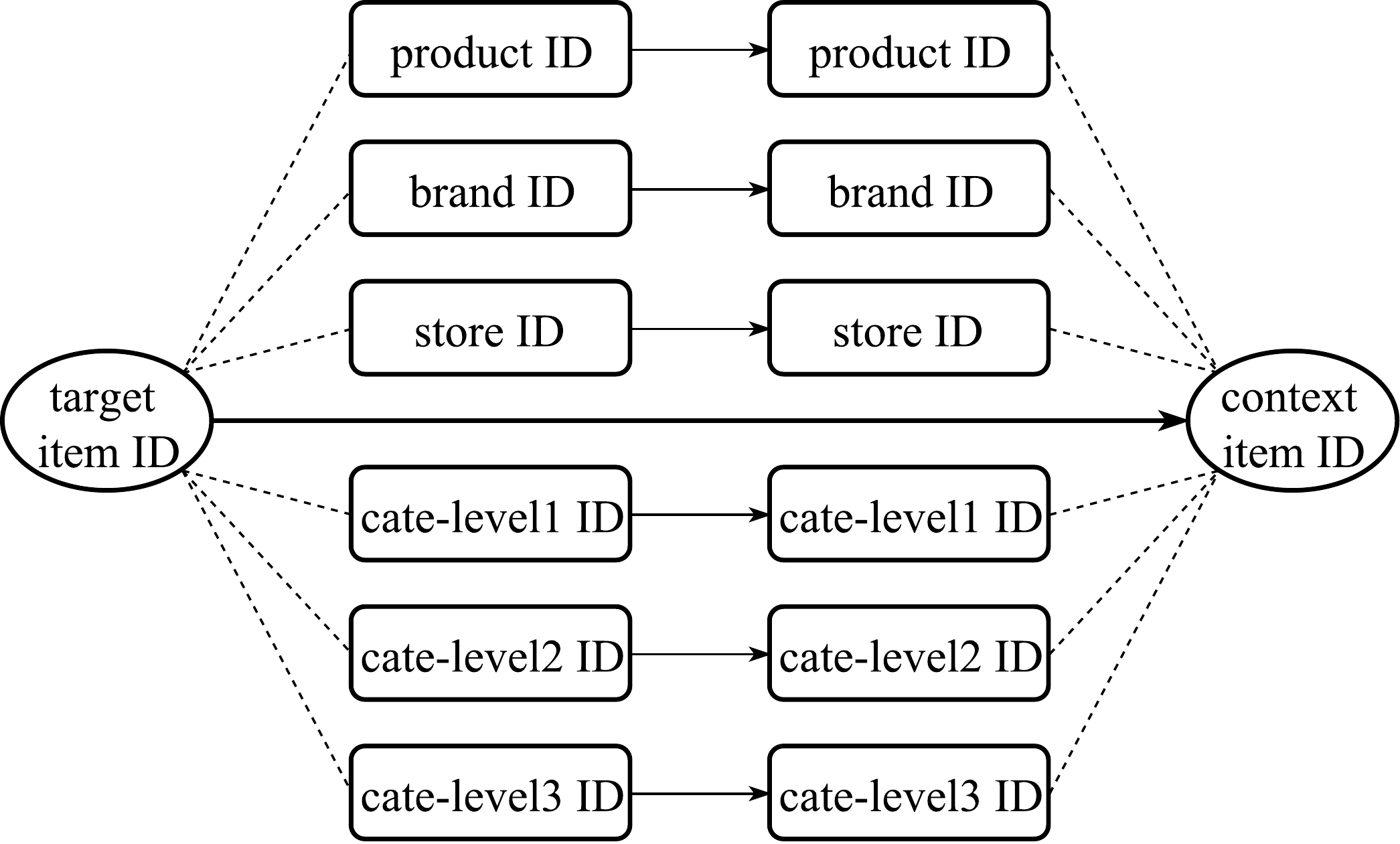}
\caption{The architecture of our jointly embedding model. }
\label{fig:jie}
\end{figure}

By exploring the structural connections between item ID and its attribute IDs, 
we propose a hierarchical embedding model to jointly 
learn low-dimensional representations for item ID as well as its attribute IDs. 
The architecture of the proposed model is shown in Figure \ref{fig:jie}, 
where the item ID is the core interactive unit and it is connected to its attribute IDs by dashed lines. 

Firstly, the co-occurrence of item IDs also implicates the co-occurrence of 
corresponding attribute IDs, which is indicated by solid arrows in Figure \ref{fig:jie}. 
Supposing there are $K (k=1,\dots, K)$ types of IDs in the first group (as mentioned in \ref{se:ia}), 
let $\text{IDs}(\text{item}_i)=[\text{id}_{1}(\text{item}_i), \dots, \text{id}_k(\text{item}_i), \dots, 
\text{id}_{K}(\text{item}_i)]$, 
where $\text{id}_{1}(\text{item}_i)$ equals $\text{item}_i$, 
and $\text{id}_{2}(\text{item}_i)$ is the product ID, 
$\text{id}_{3}(\text{item}_i)$ is the store ID and so on. 
Then we replace Eq. \ref{eq:old_match} with: 
\begin{equation}
\begin{aligned}
&p\left(\text{IDs}(\text{item}_j)|\text{IDs}(\text{item}_i)\right)\\
=&\sigma\left(\sum\limits_{k=1}^{K}{(w_{jk}{\bf e}'_{jk}})^{\rm T}(w_{ik}{\bf e}_{ik})\right)\\
&\prod\limits_{s=1}^{S}\sigma\left(-\sum\limits_{k=1}^{K}{(w_{sk}{\bf e}'_{sk}})^{\rm T}(w_{ik}{\bf e}_{ik})\right),
\label{eq:match}
\end{aligned}
\end{equation}
where ${\bf e}'_{\cdot k}\in {\bf E}'_k(\subset \mathbb{R}^{m_k\times D_k})$ 
and ${\bf e}_{\cdot k}\in {\bf E}_k(\subset \mathbb{R}^{m_k\times D_k})$. 
${\bf E}'_k$ and ${\bf E}_k$ are matrices that correspond to the context 
and target representations of type $k (k=1,\dots, K)$ respectively. 
For type $k$, $m_k$ is the dimension of its embedding vectors and $D_k$ is the size of its dictionary. 
Note that different types of IDs can be embedded into different dimensions. 
The scalar $w_{ik}$ is the weight of $\text{id}_k(\text{item}_i)$. 
Assuming each item contributes equally to $\text{id}_{k}(\text{item}_i)$ and 
$\text{id}_{k}(\text{item}_i)$ contains $V_{ik}$ different items, 
it is reasonable to let $w_{ik}$ be inversely proportional to $V_{ik}$. 
More formally, we have 
\begin{equation}
{\bf I}(x)=
\left\{\begin{matrix}
0 & ,x\text{ is False}\\ 
1 &  ,x\text{ is True}
\end{matrix}\right.,
\end{equation}
\begin{equation}
V_{ik}=\sum\limits_{j=1}^{D}{\bf I}\left(\text{id}_k(\text{item}_i) = \text{id}_k(\text{item}_j)\right),
\end{equation}
\begin{equation}
w_{ik} = \frac{1}{V_{ik}}(k=1, \dots, K). 
\end{equation}
For instance, $w_{i1}=1$ as each $\text{id}_1(\text{item}_i)$ contains exactly one item, 
and $w_{i2}=\frac{1}{10}$ if $\text{product ID}(\text{item}_i)$ contains ten different items.   

Secondly, structural connections between the item ID and attribute IDs mean constraints, 
e.g., the vectors of two item IDs should be close not only for their co-occurrence 
but also for their sharing the same product ID, store ID, brand ID or cate-level1 ID etc. 
Conversely, attribute IDs should assimilate the information contained in corresponding item IDs. 
Taking store ID as an example, the embedding vector of a specific store ID should be the proper summary of all item IDs sold in it. 
Consequently we define: 
\begin{equation}
\begin{aligned}
p(\text{item}_i|\text{IDs}(\text{item}_i))=\sigma\left(\sum\limits_{k=2}^{K}w_{ik}{\bf e}_{i1}^T{\bf M}_k{\bf e}_{ik}\right),
\label{eq:con}
\end{aligned}
\end{equation}
where ${\bf M}_k \subset \mathbb{R}^{m_1\times m_k}(k=2,\dots, K)$ is the matrix that transforms embedding vector 
${\bf e}_{i1}$ into the same dimension with embedding vector ${\bf e}_{ik}$.  
Then we maximize the following average log probability instead of Eq. \ref{eq:lp}:
\begin{equation}
\begin{aligned}
\mathcal{J}=&\frac{1}{N}\sum\limits_{n=1}^{N}\left(\sum\limits_{-C\le j \le C}^{1 \le n+j \le N, j \ne 0}
\log p(\text{IDs}(\text{item}_{n+j})|\text{IDs}(\text{item}_n))\right.\\
&\left.+\alpha\log p(\text{item}_n|\text{IDs}(\text{item}_n))-\beta\sum\limits_{k=1}^K||{\bf M}_k||_2\right),
\label{eq:jie}
\end{aligned}
\end{equation}
where $\alpha$ is the strength of constraints among IDs 
and $\beta$ is the strength of L2 regularization on transformation matrices. 

Our approach embeds the item ID and its attribute IDs into one semantic space, 
which is a useful property for deploying and transferring 
these representations in real world scenarios. 
As the properties of item ID and its attribute IDs remain stable for a relative long time, 
the jointly embedding model and the learned representations are updated weekly in our work.  

\subsection{Embedding User IDs}
\label{se:uie}
The user preferences can be reflected from their interactive sequences of item IDs, 
and thus it is reasonable to represent the user IDs by aggregating embedding vectors 
of the interactive item IDs. There are many methods to aggregate item embedding vectors, 
e.g. Average, RNN etc. \cite{okura2017embedding}, 
and Average is chosen in our work. Since the preferences of users in Hema change quickly, 
the embedding vectors of user IDs should be updated frequently (such as updated daily), 
to immediately reflect the latest preference. 
Unlike the RNN model, which needs the training procedure and is very computational expensive, 
Average is able to learn and update representations in short periods of time.

For user $u\in U$, let $S_u = [\text{item}_1, \dots,  \text{item}_t, \dots , \text{item}_T]$ 
denote the interactive sequence, where the recent $T$ item IDs are arranged in reverse chronological order. 
We construct the embedding vector for user $u$ by: 
\begin{equation}
\label{eq:ue}
\text{Embedding}(u)=\frac{1}{T}\sum_{t=1}^{T}e_t, 
\end{equation}
where $e_t$ is the embedding vector of $\text{item}_t$. 

\subsection{Model Learning}
\label{section:model_learning}
Optimizing the jointly embedding model is equivalent to 
maximizing the log-likelihood given by Eq. \ref{eq:jie}, 
which is approximated by log-uniform negative-sampling (as shown in Eq. \ref{eq:match}). 
To solve the optimization problem, we first initialize all trainable parameters using ``Xavier'' initialization \cite{glorot2010understanding}. 
After that we apply the Stochastic Gradient Descent (SGD) algorithm to $\mathcal{J}$ with shuffled mini-batches.  
The parameters are updated through back propagation 
(see \cite{goodfellow2016deep} for its principle) with Adam rule \cite{kinga2015method}. 
To exploit the parallelism of operations for speeding up, 
we train our neural network on the NVIDIA-GPU with Tensorflow \cite{abadi2016tensorflow}.

The hyper-parameters in our model are set as follows: 
the length of the context window is $C=4$; the number of negative samples is $S=2$;
the embedding dimensions are $[m_1, m_2, m_3, m_4, m_5, m_6, \\m_7]=[100, 100, 10, 20,10, 10, 20]$; 
the strength of constraints is $\alpha=1.0$; 
the strength of L2 regularization on transformations is $\beta=0.01$; 
the batch size is 128 and the neural network is trained for 5 epochs.

\section{Deploying IDs Representation}
The low-dimensional representations learned by our approach can be 
effectively deployed in many applications in E-commerce. 
In this section, we give four real-world examples in Hema App.

\subsection{Measuring Items Similarity}
\label{section:item_similarity}
Computing item-item relationships is a key building block in modern recommendation systems. 
These relationships are extensively used in many recommendation tasks. 
One classic task is ``People also like'', in which similar items are recommended 
to users based on the item they are viewing. 
Most online stores, e.g., Amazon, Taobao, Netfix and iTunes store etc., provide similar recommendation lists.  
Another classic task is to prepare for user-item recommendations, 
where Click Through Rates (CTR) are calculated between users and items (as mentioned in Section \ref{se:hema}). 
One typical choice of trigger is visited items and the candidate set is constructed based on item-item similarities. 
Given user $u$, we take the items interacted by $u$ in recent past as the seed set (denoted by SEED$(u)$). 
For each $\text{seed}_i \in \text{SEED}(u)$, we take into account its top-N similar items and the candidate set 
for user $u$ is: 
\begin{equation}
\begin{aligned}
\text{candidate}(u)=\bigcup_{\text{seed}_i \in \text{SEED}(u)}(\text{top-N similar items of seed}_i).
\end{aligned}
\end{equation} 

The item-based Collaborative Filtering (CF) \cite{sarwar2001item, linden2003amazon} 
is a famous method of calculating item-item similarities. It calculates similarity scores using user-item interactions. 
However, user-item relationships are not always available. A large proportion of interactions in online shopping 
has no explicit user identification. Instead, the session identification is available in most situations. 
We will see that measuring item similarities by cosine distances among embedding vectors has a higher recall score 
than the classical item-based CF algorithm. In practice, by integrating the new similarities into original scores 
calculated by CF algorithm, the performance of online recommendation system is significantly improved. 

\subsection{Transferring from Seen Items to Unseen Items}
New items cause the cold-start problem, which means item IDs with no historical records are invisible 
to recommendation systems \cite{schein2002methods}. 
Many existing methods cannot process new items, including item-based CF and original item2vec \cite{barkan2016item2vec} etc. 
Several content-based approaches are proposed to address this problem, 
such as image-based \cite{mcauley2015image} and text-based \cite{kui2017deep} recommendations. 
In this paper, we propose a new method to relieve the cold-start problem on unseen items, 
where an approximate embedding vector is constructed for the new item ID. 
It is facilitated by the measurability of relationships among heterogeneous IDs. 

The basic idea is that IDs connected to the new item ID usually have historical records. 
For instance, given a new item, its corresponding product is likely sold elsewhere 
and the corresponding store has already sold other items.   
Therefore we can construct an approximate embedding vector for the new item ID 
from the embedding vectors of IDs connected to it. 
Since $\sigma$ is a monotonically increasing function, we can derive Eq. \ref{eq:con} as following: 
\begin{equation}
\begin{aligned}
&p(\text{item}_i|\text{IDs}(\text{item}_i))\\
&=\sigma\left(\sum\limits_{k=2}^{K}w_{ik}{\bf e}_{i1}^T{\bf M}_k{\bf e}_{ik}\right)\\
&\propto \sum\limits_{k=2}^{K}w_{ik}{\bf e}_{i1}^T{\bf M}_k{\bf e}_{ik}\\
&={\bf e}_{i1}^T\left(\sum\limits_{k=2}^{K}w_{ik}{\bf M}_k{\bf e}_{ik}\right).
\end{aligned}
\end{equation}

Maximizing the log-likelihood given by Equation 12 leads to $p(\text{item}_i|\text{IDs}(\text{item}_i))\to 1$ 
and thus it is reasonable to approximate ${\bf e}_{i1}$ according to:
\begin{equation}
\begin{aligned}
&p(\text{item}_i|\text{IDs}(\text{item}_i))\to 1\\
& \Rightarrow {\bf e}_{i1}^T\left(\sum\limits_{k=2}^{K}w_{ik}{\bf M}_k{\bf e}_{ik}\right)\text{ is relatively large}\\
& \Rightarrow {\bf e}_{i1} \approx \sum\limits_{k=2}^{K}w_{ik}{\bf e}_{ik}^T{\bf M}_k^T.
\end{aligned}
\end{equation}
We here assume every connected $\text{id}_k(k=2, \dots, K)$ has historical records for simplicity. 
In practice, we only take into account the $\text{id}_k$ which has historical records. 
Actually, the more connected IDs are taken into account, the better approximation we can get.  
The experimental results show that the constructed vector 
$\sum\limits_{k=2}^{K}w_{ik}{\bf e}_{ik}^T{\bf M}_k^T$ is very encouraging.

\subsection{Transferring across Different Domains}
For emerging platforms like Hema, a high proportion of users are new customers, 
and thus the personalized recommendation is very challenging. 
In this work, we transfer the preferences of users on long existing platforms (source domain), 
onto the emerging platform (target domain) to overcome this challenge. 
Since Taobao covers the majority of E-commerce users in China, we select it as the source domain. 

\begin{figure}[htbp!]
\centering
\includegraphics[width=7.5cm]{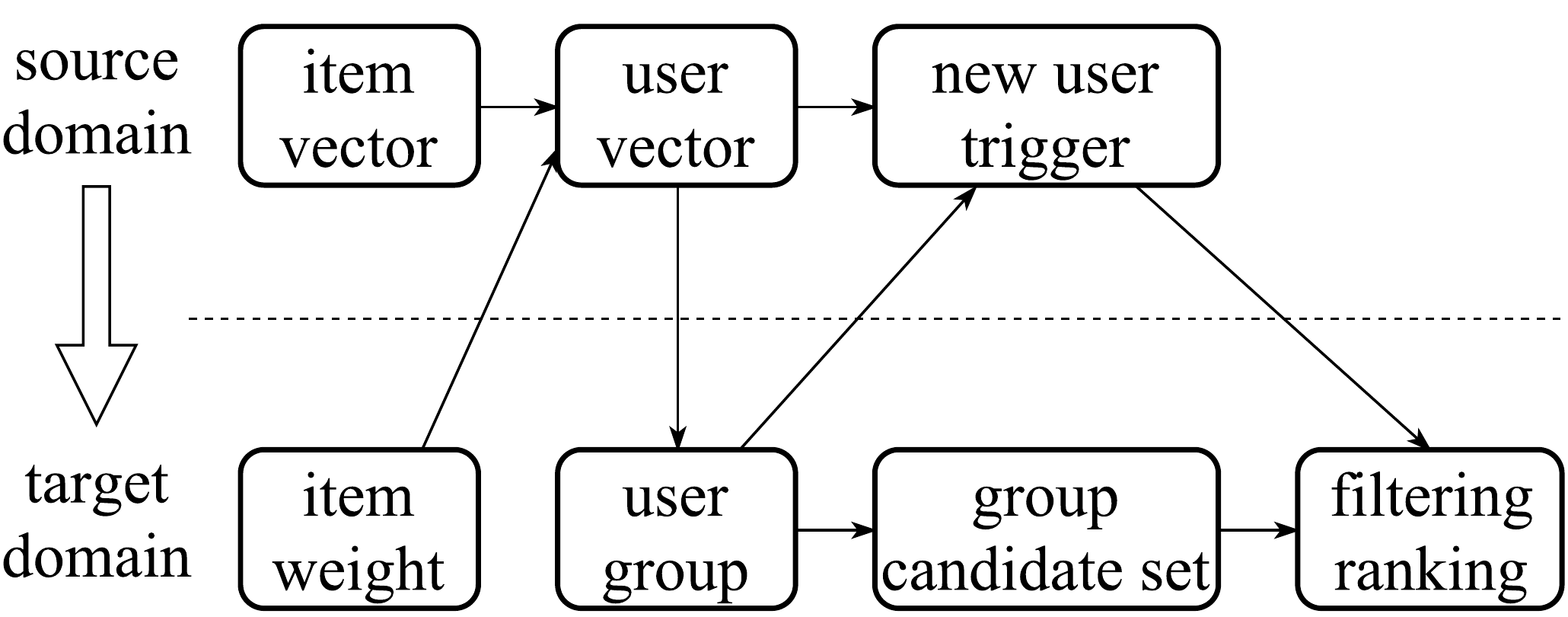}
\caption{The process of transferring user vectors across different domains.}
\label{fig:dt}
\end{figure}

Let $U^s$ and $U^t$ denote the user set in source domain and target domain respectively, 
and let $U^i$ denote their intersection $U^s\bigcap U^t$. 
The overall process is shown in Figure \ref{fig:dt}. 
Firstly, user vectors for $U^s$ are calculated via aggregating embedding vectors of the interactive item IDs in Taobao, 
which is detailed in Section \ref{se:uie}.
Secondly, based on the similarities measured by embedding vectors, users in $U^i$ are clustered into 1000 groups by k-means. 
Thirdly, for each group, the top $N$ most popular Hema items within it are selected as the candidate set. 
Fourthly, the new coming user in $U^s\setminus U^t$ is assigned to the most similar group according to the similarities 
between the user vector and the group centers. 
Finally, taking assigned group as the trigger, 
corresponding candidate set is recommended to the new user after being filtered and ranked accordingly. 

There are two alternative aggregating manners: naive average and weighted average. 
In the weighted average, different interactive items have different weights: 
\begin{equation}
\begin{aligned}
w_t &= \sum_{u'\in U^i} \text{IA}(u', \text{item}_t), \\
{\bf IA}(x) & =
\left\{\begin{matrix}
0, &\text{ $u'$ has no interaction on $\text{item}_t$}\\ 
A_j, & \text{$u'$ has interaction $j$ on $\text{item}_t$}
\end{matrix}\right..
\end{aligned}
\end{equation}
In this way, the items that are similar to items in the target domain have higher weights. 
The $A_j$ for different types of interactions are tuned by online experiments, 
e.g. purchase is more heavily weighted than click. 
Subsequently for user $u\in U^s$, the embedding vector in Eq. \ref{eq:ue} is extended to:
\begin{equation}
\text{Embedding}(u)=\frac{\sum_{t=1}^{T}w_te_t}{\sum_{t=1}^{T}w_t}.  
\end{equation}

We will see that even naive average brings significant improvement, 
and weighted average further amplifies the improvement.  

\subsection{Transferring across Different Tasks}
Sales forecast is very helpful for making informed business decisions in E-commerce. 
It can help to manage the workforce, cash flow and resources, such as optimizing the delivery process. 
We here consider the problem of forecasting the delivery demand of each store per 30 minutes in the next day. 
The forecast results can guide us to pre-order appropriate number of delivery staff.  

The value of sales forecast depends on its accuracy and inaccurate forecasts deteriorate decision efficiency.
Ordering too many delivery staff leads to manpower waste 
and ordering insufficient delivery staff causes that orders cannot be delivered in time.
Typical forecast techniques take historical sales data as input. 
However, the store ID is also very useful since different stores follow different patterns, 
and the forecasting model should be able to distinguish them. 
One classical way to consider store ID is one-hot encoding, which treats store IDs as atomic units. 
Although the simplicity and robustness make it very popular in statistical models, 
it has several limitations, as mentioned in Section \ref{section:intro}. 

In this paper, we use the embedding vectors learned by our model to represent store IDs. 
To forecast the delivery demand, we take both historical sales data and embedding vectors as the input. 
Following the input layer, several full connections with activation function are used to extract high order representations. 
Formally, let ${\bf h}_0$ denote the input layer, then the output of each hidden layer $i(i=1, \dots, N)$ is: 
\begin{equation}
{\bf h}_i = {\text a}({\bf W}_i{\bf h}_{i-1}+{\bf b}_i). 
\end{equation}
The element-wise activation function ${\text a}(\cdot)$ makes the network capable of learning non-linear functions. 
In practice, we choose the ReLU (Rectified Linear Unit defined as max(0, x)) due to its simplicity and computing efficiency. 
${\bf W}_i$ is the affine transformation of hidden layer $i$ and ${\bf b}_i$ is the corresponding bias term. 
After all hidden layers, we use the linear regression to forecast the delivery demand $\hat{y}_i$: 
\begin{equation}
\hat{y}_i={\bf w}^T{\bf h}_N+b.
\end{equation}

We train the model to minimize the Mean Absolute Error (MAE) and the learning details are similar to Section \ref{section:model_learning}. 
We will see that these embedding vectors are more effective in representing stores than the one-hot encoding.

\section{Experiments}
In this section, we present the empirical evaluation of our approach in real-word scenarios of Hema App. 

\subsection{Measuring Items Similarity}
\label{se:mis}
The item-item similarities are extensively used in many recommendation tasks, 
including ``People also like'' and CTR prediction etc. 
The similarity between two item IDs can be measured by the cosine similarity between their vectors:
\begin{equation}
\begin{aligned}
\text{sim}(\text{item}_i, \text{item}_j)&=\cos({\bf v}_i, {\bf v}_j)=\frac{{\bf v}_i^T{\bf v}_j}{||{\bf v}_i||_2 \cdot ||{\bf v}_j||_2}.
\end{aligned}
\end{equation}

When training our model, historical interaction data in the past 14 days are used. 
We construct the candidate set for each user $u_i$ as described in Section \ref{section:item_similarity}, 
and measure the offline performance of different methods by the click recall@top-N of the generated 
candidate set in the next day:
\begin{equation}
\text{recall@top-N}=\frac{\sum_{u_i}\text{\#\{hits in top-N\}}_{u_i}}{\sum_{u_i}\text{\#\{total clicks\}}_{u_i}}.
\end{equation}
Given a user $u_i$, the term $\text{\#\{hits in top-N\}}_{u_i}$ denotes the number of clicked items hit by candidate set 
and the term $\text{\#\{total clicks\}}_{u_i}$ denotes the total number of clicked items.

\begin{table*}[!htbp]
\caption{The click recall@top-N of all methods (higher is better).}
\begin{center}
\begin{tabular}{c|c|c c c c c c c c c c c}
\toprule
& top-N & 10 & 20 & 30 & 40 & 50 & 60 & 70 & 80 & 90 & 100 & 1000 \\
\hline
\multirow{2}{*}{weekday}
& CF & 2.46\% & 4.46\% & 6.07\% & 7.44\% & 8.66\% & 9.82\% & 10.88\% & 11.82\% & 12.69\% & 13.49\% & 29.83\% \\
& ITEM2VEC & 4.72\% & 7.46\% & 9.43\% & 11.00\% & 12.35\% & 13.53\% & 14.16\% & 14.73\% & 15.26\% & 15.74\% & 30.35\% \\
\hline
\hline
\multirow{2}{*}{weekend}
& CF & 3.44\% & 6.18\% & 8.46\% & 10.39\% & 12.12\% & 13.65\% & 15.05\% & 16.32\% & 17.49\% & 18.57\% & 42.33\% \\
& ITEM2VEC & 6.49\% & 10.42\% & 13.29\% & 15.58\% & 17.48\% & 19.04\% & 19.81\% & 20.55\% & 21.23\% & 21.88\% & 43.34\% \\
\bottomrule
\end{tabular}
\end{center}
\label{tb:recall_top_N}
\end{table*}

\begin{table*}[!htbp]
\caption{The click recall@top-1000 of all methods at different popularity levels (higher is better).}
\begin{center}
\begin{tabular}{c|c c c c c c c c c c}
\toprule
popular-level & 1 & 2 & 3 & 4 & 5 & 6 & 7 & 8 & 9 & 10 \\
\hline
CF & 22.67\% & 31.53\% & 36.81\% & 39.85\% & 43.17\% & 46.72\% & 47.22\% & 47.50\% & 45.85\% & 58.27\% \\
ITEM2VEC & 25.13\% & 40.44\% & 45.02\% & 47.14\% & 49.33\% & 51.34\% & 49.70\% & 49.72\% & 47.99\% & 48.97\% \\
\bottomrule
\end{tabular}
\end{center}
\label{tb:recall_level}
\end{table*}

\begin{table*}[!htbp]
\caption{The click recall@top-N of baselines and constructed vectors (higher is better).}
\begin{center}
\begin{tabular}{c|c|c c c c c c c c c c c c}
\toprule
& top-N & 10 & 20 & 30 & 40 & 50 & 60 & 70 & 80 & 90 & 100 & 1000 \\
\hline
\multirow{3}{*}{weekday}
& RANDOM & 0.01\% & 0.01\% & 0.02\% & 0.03\% & 0.04\% & 0.04\% & 0.05\% & 0.06\% & 0.07\% & 0.07\% & 0.53\% \\
& HOT          &  1.60\% & 2.46\% & 3.19\% & 3.77\% & 4.36\% & 4.83\% & 5.39\% & 5.86\% & 6.39\% & 6.97\% & 27.67\% \\
& NEW2VEC & 4.56\% & 7.05\% & 8.73\% & 9.94\% & 10.88\% & 11.63\% & 11.75\% & 11.87\% & 11.99\% & 12.11\% & 16.95\% \\
\hline
\hline
\multirow{3}{*}{weekend}
& RANDOM & 0.00\% & 0.01\% & 0.02\% & 0.03\% & 0.04\% & 0.05\% & 0.06\% & 0.07\% & 0.08\% & 0.08\% & 0.70\% \\
& HOT          & 1.69\% & 2.51\% & 3.19\% & 3.79\% &4.35\% & 4.83\% & 5.42\% & 6.00\% & 6.53\% & 7.13\% & 27.75\% \\
& NEW2VEC & 6.26\% & 9.86\% & 12.27\% & 14.02\% & 15.33\% & 16.37\% & 16.47\% & 16.56\% & 16.66\% & 16.75\% & 21.89\% \\
\bottomrule
\end{tabular}
\end{center}
\label{tb:reconstruct}
\end{table*}

Table \ref{tb:recall_top_N} shows the experimental results about click recall@top-N of two methods. 
As we can see, the low-dimensional embedding methods outperform item-based CF method \cite{sarwar2001item, linden2003amazon} significantly, 
especially when N is small. 
The reason is that item-based CF needs explicit user-item relationships and they are not always available. 
Instead of explicit user identification, the available information is session identification and skip-gram 
on the interaction sequence can take advantage of all data. 
A noticeable phenomenon is that the recall@top-N over the weekend is higher than the weekday, 
which is consistent with the experiences in our daily life. 

To get a better understanding about the improvements, 
we uniformly divide all items from the weekend dataset into 10 popularity levels according to their frequencies of interactions. 
After that we calculate the click recall@top-1000 at each level and 
show the experimental results in Table \ref{tb:recall_level}. 
Although unpopular items have few explicit user-item relationships, 
they appear in a considerable number of interaction sequences. 
Therefore embedding methods achieve better results, 
whereas the item-based CF method cannot deal with unpopular items very well. 

In the online recommendation system, by integrating the new similarities into original scores calculated by CF algorithm, 
the final recall increases by 24.0\%.

\subsection{Transferring from Seen Items to Unseen Items}
New items cause cold-start problem, and many methods cannot process them.
We attempt to relieve the cold-start problem by constructing an approximate embedding vector for the new item ID. 
Since item-based CF cannot perceive item IDs with no historical records, 
we compare our approximation to the baselines where candidate sets consist of random or hot item IDs. 
We measure the performance of different methods by the click recall@top-N of the generated candidate set in the next day. 

Table \ref{tb:reconstruct} shows the experimental results about click recall@top-N of baselines 
and our method. As we can see, the performance of constructed vectors is competitive. 
Considering that the classic item-based CF cannot deal with new item IDs at all, our approach is very encouraging. 
The recall@top-N on the weekend is higher than the weekday, which is similar to Section \ref{se:mis}. 

On closer inspection, when N is less than 50 the click recall is comparable to 
that using item ID embedding vectors directly. 
When N is beyond 50, few improvements can be achieved. 
That reveals the limitation of our approach: the approximation is accurate when N is small 
and the approximation becomes inaccurate while N increases. In practice, 
the problem brought by this limitation is negligible 
in that the number of items we can recommend to a user is very limited. 

\subsection{Transferring across Different Domains}
Since new users have no historical records, it is impossible to evaluate different methods offline. 
Instead, we conduct an A/B test over one week and three methods are compared: 
\begin{itemize}
\item {\bf Hot.} The baseline is recommending the same list containing hot items to each new user. 
\item {\bf Naive average.} Through transferring the preferences of users on Taobao to Hema, 
personalized recommendation is provided to each new user. In this method, the user vector 
is generated by naive average. 
\item {\bf Weighted average.} When generating user vectors, different weights are here 
assigned to different interactive items and the weighted average is adopted. 
\end{itemize}

Performances of different methods are evaluated by Pay-Per-Impression (PPM). 
Compared to the baseline, the naive average increased PPM by 71.4\% 
and the weighted average increased PPM by 141.8\%. 
The significant improvements demonstrate that user embedding vectors learned by our approach 
can be effectively transferred across different domains to overcome the cold-start problem 
brought by new users.

\subsection{Transferring across Different Tasks}
We here consider the problem of forecasting the delivery demand of each store per 30 minutes on the next day. 
The forecast results can guide us to pre-order appropriate number of delivery staff. 
We compare our embedding vectors against several baselines: 
\begin{itemize}
\item {\bf HISTORY.} The typical forecast techniques take historical sales data as the input. 
For every 30-minute interval of each store, the input contains the delivery demands in the same time interval 
on the recent 7 days. 
\item {\bf HISTORY with ONE-HOT.} The store ID is also very useful since different stores follow different patterns. 
One classical way to consider store ID is one-hot encoding, which treats store IDs as atomic units. 
Besides the historical data, we here also take the one-hot encoding as part of the input. 
\item {\bf HISTORY with VEC.} One-hot encoding has several limitations, such as sparsity and similarity immeasurability etc. 
We represent the store ID using the embedding vector learned by our embedding model 
and take it as part of the input instead of one-hot encoding. 
\end{itemize}

There are five full connection layers of size 128 in every alternative methods. 
We measure the performance of different methods by the Relative Mean Absolute Error (RMAE): 
\begin{equation}
\text{RMAE}=\frac{\sum_{i=1}^{N}|y_i-\hat{y}_i|}{\sum_{i=1}^{N}y_i},
\end{equation}
where $y_i$ is the true delivery demand and $\hat{y}_i$ is the forecasted delivery demand. 
Note that it is different from the Mean Absolute Percentage Error (MAPE). 
RMAE actually is an equivalent variety of MAE since $\sum_{i=1}^{N}y_i$ is a constant for a certain dataset. 

\begin{table}[!thbp]
\caption{The RMAE scores of different methods in forecasting delivery demand (lower is better).}
\begin{center}
\begin{tabular}{c|c c c}
\toprule
dataset & day 1 & day 2 & day 3 \\
\hline
HISTORY & 43.23\% & 40.75\% & 34.26\% \\
HISTORY with ONE-HOT & 42.57\% & 39.00\% & 34.57\% \\
HISTORY with VEC & {40.95\%} & {33.75\%} & {33.02\%} \\
\bottomrule
\end{tabular}
\end{center}
\label{tb:forecast}
\end{table}

The RMAE scores of different methods in forecasting delivery demand are shown in Table \ref{tb:forecast}. 
We repeat the experiment on 3 different days to confirm the improvement. 
As we can see, the store IDs are very useful and even one-hot encoding can bring improvements. 
However, one-hot encoding is plagued by its sparsity and distance unavailability. 
The embedding vectors of store IDs learned by our embedding model can overcome those limitations. 
By taking full advantage of the information contained in store IDs, 
our model achieves much more improvements.

\section{Related Work}
Many machine intelligence techniques, such as recommendation \cite{bobadilla2013recommender} 
and forecast \cite{beheshti2015survey} etc., 
are developed to handle the dynamic and complex business environment in E-commerce.  
The performance of these intelligence techniques is heavily dependent on the data representation 
and thus representation learning has been a very popular topic for a while \cite{bengio2013representation}. 
Recently, many works focus on a specific family of representation learning methods, 
namely neural networks (or deep learning). It is inspired by the nervous system and consists of 
multiple non-linear transformations. Representation learning with neural networks has brought 
remarkable successes to both academia and 
industry \cite{krizhevsky2012imagenet, graves2013speech, blunsom2014convolutional}. 

The unordered discrete data (or called IDs) is one of the most important types of data in many scenarios. 
The word in Natural Language Processing (NLP) is a typical example of IDs. 
Traditional techniques treat them as atomic units and represent them as indices or one-hot encodings. 
This simple and robust choice has many limitations, including sparsity and distance unavailability. 
To overcome these limitations, low-dimensional distributed representations are proposed, 
which can be traced back to the classical neural network language model \cite{bengio2003a}. 
They jointly learn the word embedding vector and the statistical language model 
with a linear projection layer and a non-linear hidden layer. 
It was later shown that word embedding vectors significantly improve 
and simplify many NLP applications \cite{collobert2008unified, collobert2011natural}. 
However, the original neural network language model is very computationally expensive 
since it involves many dense matrix multiplications. 
Mikolov et al. introduced the skip-gram model \cite{mikolov2013efficient} (also well-known as word2vec), 
which predicts the surrounding context words given the target word to avoid many dense matrix multiplications. 
Although word2vec is very efficient to learn high-quality embedding vectors of words, 
Mikolov et al. used the negative-sampling to approximate full softmax 
to reduce the computation further \cite{mikolov2013distributed}. 
Their works have been followed by many others. Le et al. developed an algorithm 
to represent each document with a vector by training a language model 
to predict words in the document \cite{le2014distributed}. 
Li et al. \cite{li2016joint} and Yamada et al. \cite{yamada2016joint} both jointly map words and
entities (from the knowledge base) into the same semantic vector space. 

Embedding techniques have drawn many attentions from various domains beyond the original NLP domain. 
Perozzi et al. \cite{perozzi2014deepwalk}, Tang et al. \cite{tang2015line} and Grover et al. \cite{grover2016node2vec} 
established an analogy for networks by representing a network as a ``document''. 
They sample sequences of nodes from the underlying network and turn a network into an ordered sequence of nodes. 
In E-commerce, Oren et al. proposed item2vec \cite{barkan2016item2vec} to embed item IDs into 
the low-dimensional representation by modeling item ID co-occurrence in user's interactive sequences. 
Sun et al. \cite{sun2017mrlr} jointly learned user ID and item ID embedding vectors 
from a multi-level item category organization for personalized ranking in recommendation. 
Ai et al. \cite{ai2017learning} jointly learned distributed representations 
for query ID, item ID and user ID for personalized product search. 
Benjamin Paul et al. \cite{chamberlain2017customer} modeled customer life time value based on user embedding. 

\section{Conclusion}
The unordered discrete ID is one of the most important types of data in many scenarios. 
In this paper, we propose an embedding based framework to learn and transfer 
low-dimensional representations for all types of IDs in E-commerce, 
including user ID, item ID, product ID, store ID, brand ID and category ID etc. 
Besides co-occurrence of item IDs in user's implicit interactions, 
we also consider the structural connections among different types of IDs.  
Then all types of IDs are embedded into one semantic space and 
these low-dimensional representations can be effectively used in many applications. 
We deploy and evaluate the proposed approach in several real-world scenarios of Hema App, 
and the results validate its effectiveness. 
We plan to continue improving our approach and extend it to many other applications, 
such as search engines, advertisements and so on. 

\begin{acks}
The authors would like to thank Chi Zhang, Zhibin Wang, Xiuyu Sun, Hao Li for discussions and supports. 
\end{acks}

\bibliographystyle{ACM-Reference-Format}
\bibliography{bibliography} 

\end{document}